\documentclass[runningheads]{llncs}
\usepackage{graphicx}
\usepackage{amsmath,amssymb,amsfonts}
\usepackage{multirow}
\usepackage{changes}
\usepackage[hidelinks]{hyperref}
\def\x{{\mathbf x}}
\def\p{{\mathbf p}}

\begin{document}
\title{Bounding Box Tightness Prior for Weakly Supervised Image Segmentation}
%
%
\author{Juan Wang\inst{1}\orcidID{0000-0003-3124-9901} \and
Bin Xia\inst{2}\orcidID{0000-0002-0340-8082}}
\authorrunning{J. Wang and B. Xia}
%
\institute{Delta Micro Technology, Inc., Laguna Hills, CA 92653 USA 
\email{wangjuan313@gmail.com}\\
\and
Shenzhen SiBright Co. Ltd., Shenzhen, Guangdong 518052 China\\
\email{b.xia@sibionics.com}}
\maketitle              
\begin{abstract}
This paper presents a weakly supervised image segmentation method that adopts tight bounding box annotations. It proposes generalized multiple instance learning (MIL) and smooth maximum approximation to integrate the bounding box tightness prior into the deep neural network in an end-to-end manner. In generalized MIL, positive bags are defined by parallel crossing lines with a set of different angles, and negative bags are defined as individual pixels outside of any bounding boxes. Two variants of smooth maximum approximation, i.e., $\alpha$-softmax function and $\alpha$-quasimax function, are exploited to conquer the numeral instability introduced by maximum function of bag prediction. The proposed approach was evaluated on two pubic medical datasets using Dice coefficient. The results demonstrate that it outperforms the state-of-the-art methods. The codes are available at \url{https://github.com/wangjuan313/wsis-boundingbox}.

\end{abstract}

\section{Introduction}

In recent years, image segmentation has been made great progress with the development of deep neural networks in a fully-supervised manner \cite{ronneberger2015u,long2015fully,chen2018encoder}. However, collecting large-scale training set with precise pixel-wise annotation is considerably labor-intensive and expensive. To tackle this issue, there have been great interests in the development of weakly supervised image segmentation. All kinds of supervisions have been considered, including image-level annotations \cite{ahn2019weakly,wang2020self}, scribbles \cite{lin2016scribblesup}, bounding boxes \cite{rajchl2016deepcut,kervadec2020bounding}, and points \cite{bearman2016s}. This work focuses on image segmentation by employing supervision of bounding boxes. 

In the literature, some efforts have been made to develop the weakly supervised image segmentation methods adopting the bounding box annotations. For example, Rajchl \textit{et al.} \cite{rajchl2016deepcut} developed an iterative optimization method for image segmentation, in which a neural network classifier was trained from bounding box annotations. Khoreva \textit{et al.} \cite{khoreva2017simple} employed GrabCut \cite{rother2004grabcut} and MCG proposals \cite{pont2016multiscale} to obtain pseudo label for image segmentation. 
Hsu \textit{et al.} \cite{hsu2019weakly} exploited multiple instance learning (MIL) strategy and mask R-CNN for image segmentation. 
Kervadec \textit{et al.} \cite{kervadec2020bounding} leveraged the tightness prior to a deep learning setting via imposing a set of constraints on the network outputs for image segmentation. 

In this work, we present a generalized MIL formulation and smooth maximum approximation to integrate the bounding box tightness prior into the network in an end-to-end manner. Specially, we employ parallel crossing lines with a set of different angles to obtain positive bags and use individual pixels outside of any bounding boxes as negative bags. We consider two variants of smooth maximum approximation to conquer the numeral instability introduced by maximum function of bag prediction.
The experiments on two public medical datasets demonstrate that the proposed approach outperforms the state-of-the-art methods. 

\section{Methods}
\subsection{Preliminaries}

\subsubsection{Problem description}
Suppose $I$ denotes an input image, and $Y \in \{1,2,\cdots,C\}$ is its corresponding pixel-level category label, in which $C$ is the number of categories of the objects. The image segmentation problem is to obtain the prediction of $Y$, denoted as $P$, for the input image $I$. 

In the fully supervised image segmentation setting, for an image $I$, its pixel-wise category label $Y$ is available during training. Instead, in this study we are only provided its bounding box label $B$. Suppose there are $M$ bounding boxes, then its bounding box label is $B = \{b_m, y_m\}, m = 1, 2, \cdots, M$, where the location label $b_m$ is a 4-dimensional vector representing the top left and bottom right points of the bounding box, and $y_m \in \{1,2,\cdots,C\}$ is its category label. 

\subsubsection{Deep neural network}
This study considers deep neural networks which output the pixel-wise prediction of the input image, such as Unet \cite{ronneberger2015u}, FCN \cite{long2015fully}, etc. Due to the possible overlaps of objects of different categories in images, especially in medical images, the image segmentation problem is formulated as a multi-label classification problem in this study. That is, for a location $k$ in the input image, it outputs a vector $\p_k$ with $C$ elements, one element for a category; each element is converted to the range of $[0, 1]$ by the sigmoid function.


\subsection{MIL baseline}

\subsubsection{MIL definition and bounding box tightness prior}
Multiple instance learning (MIL) is a type of supervised learning. Different from the traditional supervised learning which receives a set of training samples which are individually labeled, MIL receives a set of labeled bags, each containing many training samples. In MIL, a bag is labeled as negative if all of its samples are negative, a bag is positive if it has at least one sample which is positive. 

Tightness prior of bounding box indicates that the location label of bounding box is the smallest rectangle enclosing the whole object, thus the object must touch the four sides of its bounding box, and does not overlap with the region outside its bounding box. The crossing line of a bounding box is defined as a line with its two endpoints located on the opposite sides of the box. In an image $I$ under consideration, for an object with category $c$, any crossing line in the bounding box has at least one pixel belonging to the object in the box; any pixels outside of any bounding boxes of category $c$ do not belong to category $c$. Hence pixels on a cross line compose a positive bag for category $c$, while pixels outside of any bounding boxes of category $c$ are used for negative bags. 


\subsubsection{MIL baseline} 
\label{section:baseline_baseline}
\begin{figure}[htbp] 
	\centering
	\setlength{\tabcolsep}{2pt}
	\begin{tabular}{cc}
	\includegraphics[trim=0in 0in 0in 0in,clip,width=2in]{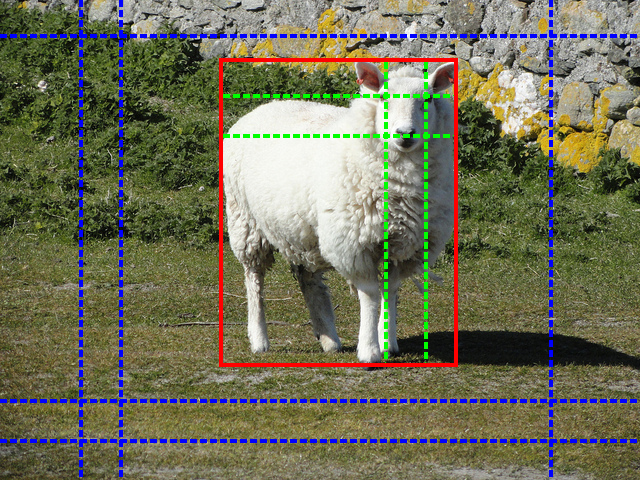} &
	\includegraphics[trim=0in 0in 0in 0in,clip,width=2in]{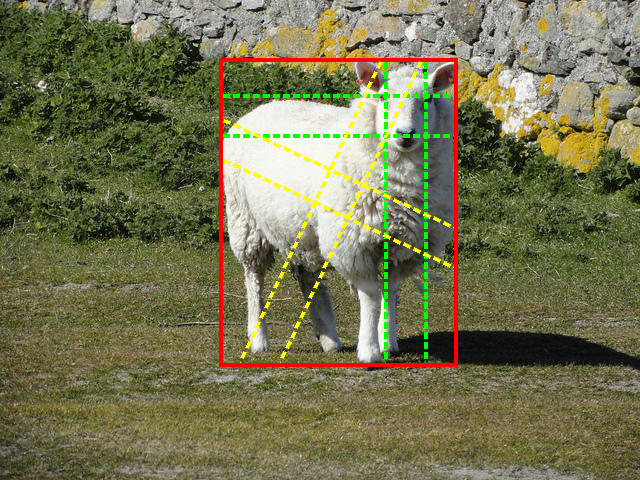} \\ 
	(a)&(b) \\
	\end{tabular}
	\caption{Demonstration of positive and negative bags, in which the red rectangle is the bounding box of the object. (a) MIL baseline. Examples of positive bags are denoted by green dashed lines, and examples of negative bags are marked by the blue dashed lines. (b) Generalized MIL. Examples of positive bags with different angles are shown ($\theta=25^\circ$ for the yellow dashed lines and $\theta=0^\circ$ for the green dashed lines). Individual pixels outside of the red box are negative bags, which are omitted here for simplicity. }
	\label{fig:mil_demonstration}
\end{figure}

For category $c$ in an image, the baseline approach simply considers all of the horizontal and vertical crossing lines inside the boxes as positive bags, and all of the horizontal and vertical crossing lines that do not overlap any bounding boxes of category $c$ as negative bags. This definition is shown in Fig.~\ref{fig:mil_demonstration}(a) and has been widely employed in the literature \cite{hsu2019weakly,kervadec2020bounding}.

To optimize the network parameters, MIL loss with two terms are considered \cite{hsu2019weakly}. For category $c$, suppose its positive and negative bags are denoted by $\mathcal{B}_c^+$ and $\mathcal{B}_c^-$, respectively, then loss $\mathcal{L}_c$ can be defined as follows:
\begin{equation}
\mathcal{L}_c = \phi_c(P; \mathcal{B}_c^+, \mathcal{B}_c^-) + \lambda \varphi_c(P)
\label{equ:mil_loss}
\end{equation}
where $\phi_c$ is the unary loss term, $\varphi_c$ is the pairwise loss term, and $\lambda$ is a constant value controlling the trade off between the unary loss and the pairwise loss.

The unary loss $\phi_c$ enforces the tightness constraint of bounding boxes on the network prediction $P$ by considering both positive bags $\mathcal{B}_c^+$ and negative bags $\mathcal{B}_c^-$. A positive bag contains at least one pixel inside the object, hence the pixel with highest prediction tends to be inside the object, thus belonging to category $c$. In contrast, no pixels in negative bags belong to any objects, hence even the pixel with highest prediction does not belong to category $c$. Based on these observations, the unary loss $\phi_c$ can be expressed as:
\begin{equation}
\phi_c = -\frac{1}{|\mathcal{B}_c^+|+|\mathcal{B}_c^-|} \left( \sum_{b \in \mathcal{B}_c^+} \log P_c(b) + \sum_{b \in \mathcal{B}_c^-} \log(1-P_c(b)) \right)
\label{equ:unary_loss}
\end{equation}
where $P_c(b) = \max_{k \in b}(p_{kc})$ is the prediction of the bag $b$ being positive for category $c$, $p_{kc}$ is the network output of the pixel location \textit{k} for category $c$, and $|\mathcal{B}|$ is the cardinality of $\mathcal{B}$. 

The unary loss is binary cross entropy loss for bag prediction $P_c(b)$, it achieves minimum when $P_c(b)=1$ for positive bags and $P_c(b)=0$ for negative bags. More importantly, during training the unary loss adaptively selects a positive sample per positive bag and a negative sample per negative bag based on the network prediction for optimization, thus yielding an adaptive sampling effect. 

However, using the unary loss alone is prone to segment merely the discriminative parts of an object. To address this issue, the pairwise loss as follows is introduced to pose the piece-wise smoothness on the network prediction.
\begin{equation}
\varphi_c = \frac{1}{|\varepsilon|} \sum_{(k,k^\prime) \in \varepsilon} \left( p_{kc} - p_{k^\prime c} \right) ^2
\end{equation}
where $\varepsilon$ is the set containing all neighboring pixel pairs, $p_{kc}$ is the network output of the pixel location \textit{k} for category $c$.

Finally, considering all $C$ categories, the MIL loss $\mathcal{L}$ is:
\begin{equation}
\mathcal{L} = \sum_{c=1}^C \mathcal{L}_c
\end{equation}

\subsection{Generalized MIL}

\subsubsection{Generalized positive bags}
For an object of height $H$ pixels and width $W$ pixels, the positive bag definition in the MIL baseline yields only $H+W$ positive bags, a much smaller number when compared with the size of the object. Hence it limits the selected positive samples during training, resulting in a bottleneck for image segmentation.

To eliminate this issue, this study proposes to generalize positive bag definition by considering all parallel crossing lines with a set of different angles. An parallel crossing line is parameterized by an angle $\theta \in (-90^\circ, 90^\circ)$ with respect to the edges of the box where its two endpoints located. For an angle $\theta$, two sets of parallel crossing lines can be obtained, one crosses up and bottom edges of the box, and the other crosses left and right edges of the box. As examples, in Fig.~\ref{fig:mil_demonstration}(b), we show positive bags of two different angles, in which those marked by yellow dashed colors have $\theta=25^\circ$, and those marked by green dashed lines are with $\theta=0^\circ$. Note the positive bag definition in MIL baseline is a special case of the generalized positive bags with $\theta=0^\circ$.  

\subsubsection{Generalized negative bags}
The similar issue also exists for the negative bag definition in the MIL baseline. To tackle this issue, for a category $c$, we propose to define each individual pixel outside of any bounding boxes of category $c$ as a negative bag. This definition greatly increases the number of negative bags, and forces the network to see every pixel outside of bounding boxes during training. 

\subsubsection{Improved unary loss}
The generalized MIL definitions above will inevitably lead to imbalance between positive and negative bags. To deal with this issue, we borrow the concept of focal loss \cite{ross2017focal} and use the improved unary loss as follows:
\begin{equation}
\phi_c = -\frac{1}{N^+} \left( \sum_{b \in \mathcal{B}_c^+} \beta \left(1-P_c(b)\right)^\gamma \log P_c(b) +  \sum_{b \in \mathcal{B}_c^-} (1-\beta)P_c(b)^\gamma \log(1-P_c(b)) \right)
\label{equ:focal_loss}
\end{equation}
where $N^+ = \max(1, |\mathcal{B}_c^+|)$, $\beta \in [0,1]$ is the weighting factor, and $\gamma \geq 0$ is the focusing parameter. The improved unary loss is focal loss for bag prediction, it achieves minimum when $P_c(b)=1$ for positive bags and $P_c(b)=0$ for negative bags.

\subsection{Smooth maximum approximation}

In the unary loss, the maximum prediction of pixels in a bag is used as bag prediction $P_c(b)$. However, the derivative $\partial P_c / \partial p_{kc}$ is discontinuous, leading to numerical instability. To solve this problem, we replace the maximum function by its smooth maximum approximation \cite{lange2014applications}. Let the maximum function be $f(\x) = \max_{i=1}^n x_i$, its two variants of smooth maximum approximation as follows are considered. 

(1) \textit{$\alpha$-softmax function:}
\begin{equation}
S_{\alpha}(\x) = \frac{\sum_{i=1}^n x_i e^{\alpha x_i}}{\sum_{i=1}^n e^{\alpha x_i}}
\end{equation}
where $\alpha>0$ is a constant. The higher the $\alpha$ value is, the closer the approximation $S_{\alpha}(\x)$ to $f(\x)$. For $\alpha \rightarrow 0$, a soft approximation of the mean function is obtained.

(2) \textit{$\alpha$-quasimax function:}
\begin{equation}
Q_{\alpha}(\x) = \frac{1}{\alpha} \log \left(\sum_{i=1}^n e^{\alpha x_i}\right) - \frac{\log n}{\alpha}
\end{equation}
where $\alpha>0$ is a constant. The higher the $\alpha$ value is, the closer the approximation $Q_{\alpha}(\x)$ to $f(\x)$. One can easily prove that $Q_{\alpha}(\x) \leq f(\x)$ always holds.

In real application, each bag usually has more than one pixel belonging to object segment. However, $\partial f / \partial x_i$ has value 0 for all but the maximum $x_i$, thus the maximum function considers only the maximum $x_i$ during optimization. In contrast, the smooth maximum approximation has $\partial S_{\alpha} / \partial x_i > 0$ and $\partial Q_{\alpha} / \partial x_i > 0$ for all $x_i$, thus it considers every $x_i$ during optimization. More importantly, in the smooth maximum approximation, large $x_i$ has much greater derivative than small $x_i$, thus eliminating the possible adverse effect of negative samples in the optimization. In the end, besides the advantage of conquering numerical instability, the smooth maximum approximation is also beneficial for performance improvement.

\section{Experiments}

\subsection{Datasets}
This study made use of two public medical datasets for performance evaluation. The first one is the prostate MR image segmentation 2012 (PROMISE12) dataset \cite{litjens2014evaluation} for prostate segmentation and the second one is the anatomical tracings of lesions after stroke (ATLAS) dataset \cite{liew2018large} for brain lesion segmentation. 


\textit{Prostate segmentation:} The PROMISE12 dataset was first developed for prostate segmentation in MICCAI 2012 grand challenge \cite{litjens2014evaluation}. It consists of the transversal T2-weighted MR images from 50 patients, including both benign and prostate cancer cases. These images were acquired at different centers with multiple MRI vendors and different scanning protocols. Same as the study in \cite{kervadec2020bounding}, the dataset was divided into two non-overlapping subsets, one with 40 patients for training and the other with 10 patients for validation. 

\textit{Brain lesion segmentation:} The ATLAS dataset is a well-known open-source dataset for brain lesion segmentation. It consists of 229 T1-weighted MR images from 220 patients. These images were acquired from different cohorts and different scanners. The annotations were done by a group of 11 experts. Same as the study in \cite{kervadec2020bounding}, the dataset was divided into two non-overlapping subsets, one with 203 images from 195 patients for training and the other with 26 images from 25 patients for validation. 

\subsection{Implementation details}
All experiments were implemented using PyTorch in this study. Image segmentation was conducted on the 2D slices of MR images. The parameters in the MIL loss (\ref{equ:mil_loss}) were set as $\lambda=10$ based on experience, and those in the improved unary loss (\ref{equ:focal_loss}) were set as $\beta=0.25$ and $\gamma=2$ according to the focal loss \cite{ross2017focal}. As indicated below, most experimental setups were set to be same as study in \cite{kervadec2020bounding} for fairness of comparison.

For the PROMISE12 dataset, a residual version of UNet \cite{ronneberger2015u} was used for segmentation \cite{kervadec2020bounding}. The models were trained with Adam optimizer \cite{kingma2014adam} with the following parameter values: batch size = 16, initial learning rate = $10^{-4}$, $\beta_1 = 0.9$, and $\beta_2 = 0.99$. To enlarge the set of images for training, an off-line data augmentation procedure \cite{kervadec2020bounding} was applied to the images in the training set as follows: 1) mirroring, 2) flipping, and 3) rotation.

For the ATLAS dataset, ENet \cite{paszke2016enet} was employed as a backbone architecture for segmentation \cite{kervadec2020bounding}. The models were trained with Adam optimizer with the following parameter values: batch size = 80, initial learning rate = $5 \times 10^{-4}$, $\beta_1 = 0.9$, and $\beta_2 = 0.99$. No augmentation was performed during training \cite{kervadec2020bounding}.


\subsection{Performance evaluation}
To measure the performance of the proposed approach, the Dice coefficient was employed, which has been applied as a standard performance metric in medical image segmentation. The Dice coefficient was calculated based on the 3D MR images by stacking the 2D predictions of the networks together.

To demonstrate the overall performance of the proposed method, we considered the baseline method in the experiments for comparison. Moreover, we further perform comparisons with state-of-the-art methods with bounding-box annotations, including deep cut \cite{rajchl2016deepcut} and global constraint \cite{kervadec2020bounding}.

\section{Results}

\subsection{Ablation study}

\subsubsection{Generalized MIL}
To demonstrate the effectiveness of the proposed generalized MIL formulation, in Table \ref{table:ablation_mil} we show the performance of the generalized MIL for the two datasets. For comparison, we also show the results of the baseline method. It can be observed that the generalized MIL approach consistently outperforms the baseline method at different angle settings. In particular, the PROMISE12 dataset got best Dice coefficient of 0.878 at $\theta_{best}=(-40^{\circ},40^{\circ},20^{\circ})$ for the generalized MIL, compared with 0.859 for the baseline method. The ATLAS dataset achieved best Dice coefficient of 0.474 at $\theta_{best}=(-60^{\circ},60^{\circ},30^{\circ})$ for the generalized MIL, much higher than 0.408 for the baseline method.

\renewcommand\arraystretch{1.3}
\begin{table}
\caption{Dice coefficients of the proposed generalized MIL for image segmentation, where $\theta=(\theta_1,\theta_2,\Delta)$ denotes evenly spaced angle values within interval $(\theta_1,\theta_2)$ with step $\Delta$. The standard deviation of Dice coefficients among different MR images are reported in the bracket. For comparison, results of the baseline method are also given.}
\centering
\label{table}
\setlength{\tabcolsep}{5pt}
\begin{tabular}{ccccc}
\hline
\hline
Method & PROMISE12 & ATLAS \\
\hline
MIL baseline & 0.859 (0.038) & 0.408 (0.249) \\
$\theta=(-40^{\circ},40^{\circ},10^{\circ})$ & 0.868 (0.031) & 0.463 (0.278) \\
$\theta=(-40^{\circ},40^{\circ},20^{\circ})$ & \textbf{0.878 (0.027)} & 0.466 (0.248) \\
$\theta=(-60^{\circ},60^{\circ},30^{\circ})$ & 0.868 (0.047) & \textbf{0.474 (0.262)} \\
\hline
\hline
\end{tabular}
\label{table:ablation_mil}
\end{table}

\subsubsection{Smooth maximum approximation}
To demonstrate the benefits of the smooth maximum approximation, in Table \ref{table:ablation_approximation} we show the performance of the MIL baseline method when the smooth maximum approximation was applied. As can be seen, for the PROMISE12 dataset, the better performance is obtained for $\alpha$-softmax function with $\alpha=4$ and 6 and for $\alpha$-quasimax function with $\alpha=6$ and 8. For the ATLAS dataset, the improved performance can also be observed for $\alpha$-softmax function with $\alpha=6$ and 8 and for $\alpha$-quasimax function with $\alpha=8$.

\renewcommand\arraystretch{1.3}
\begin{table}
\caption{Dice coefficients of the MIL baseline method when smooth maximum approximation was applied. }
\centering
\label{table}
\setlength{\tabcolsep}{5pt}
\begin{tabular}{cccc}
\hline
\hline
\multicolumn{2}{c}{Method} & PROMISE12 & ATLAS \\
\hline
\multirow{3}{*}{$\alpha$-softmax} & $\alpha = 4$ & 0.861 (0.031) & 0.401(0.246) \\
& $\alpha = 6$ & 0.861 (0.036) & 0.424(0.255) \\
& $\alpha = 8$ & 0.859 (0.030) & 0.414(0.264) \\
\hline
\multirow{3}{*}{$\alpha$-quasimax} & $\alpha = 4$ & 0.856 (0.026) & 0.405(0.246) \\
& $\alpha = 6$ & 0.873 (0.018) & 0.371(0.240) \\
& $\alpha = 8$ & 0.869 (0.024) & 0.414(0.256) \\
\hline
\hline
\end{tabular}
\label{table:ablation_approximation}
\end{table}

\subsection{Main experimental results}
In Table \ref{table:main_results} the Dice coefficients of the proposed approach are given, in which $\alpha=4,6$, and 8 are considered in smooth maximum approximation functions and those with highest dice coefficient are reported. For comparison, the full supervision results are also shown in Table \ref{table:main_results}. As can be seen, the PROMISE12 dataset gets Dice coefficient 0.878 for $\alpha$-softmax function and 0.880 for $\alpha$-quasimax function, which are same as or higher than the results in the ablation study. More importantly, these values are close to 0.894 obtained by the full supervision. The similar trends are observed for the ATLAS dataset.

\renewcommand\arraystretch{1.3}
\begin{table}
\caption{Comparison of Dice coefficients for different methods.}
\centering
\label{table}
\setlength{\tabcolsep}{5pt}
\begin{tabular}{ccccc}
\hline
\hline
Method & PROMISE12 & ATLAS \\
\hline
Full supervision & 0.894 (0.021) & 0.512 (0.292) \\
$\theta_{best}$ + $\alpha$-softmax & \textbf{0.878 (0.031)} & \textbf{0.494 (0.236)} \\
$\theta_{best}$ + $\alpha$-quasimax & \textbf{0.880 (0.024)} & \textbf{0.488 (0.240)} \\
Deep cut \cite{rajchl2016deepcut} & 0.827 (0.085) & 0.375 (0.246) \\
Global constraint \cite{kervadec2020bounding} & 0.835 (0.032) & 0.474 (0.245) \\
\hline
\hline
\end{tabular}
\label{table:main_results}
\end{table}

Furthermore, we also show the Dice coefficients of two state-of-the-art methods in Table \ref{table:main_results}. The PROMISE12 dataset gets Dice coefficient 0.827 for deep cut and 0.835 for global constraint, both of which are much lower than those of the proposed approach. Similarly, the proposed approach also achieves higher Dice coefficients compared with these two methods for the ATLAS dataset. 

\begin{figure}[htbp] 
	\centering
	\setlength{\tabcolsep}{2pt}
	\begin{tabular}{cccc}
	\includegraphics[trim=0in 0in 0in 0in,clip,width=1in]{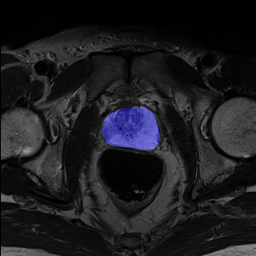} &
	\includegraphics[trim=0in 0in 0in 0in,clip,width=1in]{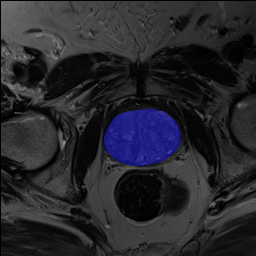} &
	\includegraphics[trim=0.4in 0in 0.25in 0in,clip,width=1in]{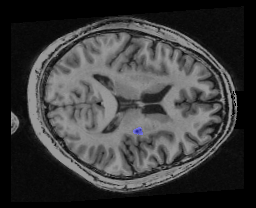} &
	\includegraphics[trim=0.4in 0in 0.25in 0in,clip,width=1in]{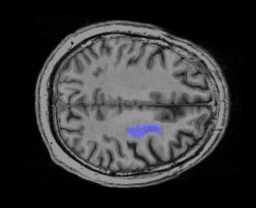} \\ 
	\includegraphics[trim=0in 0in 0in 0in,clip,width=1in]{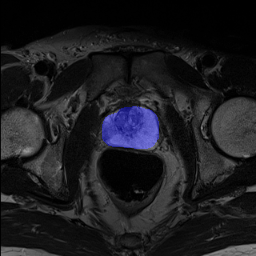} &
	\includegraphics[trim=0in 0in 0in 0in,clip,width=1in]{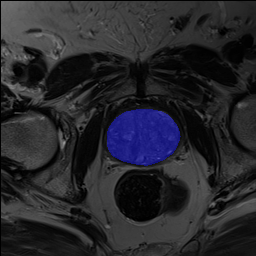} &
	\includegraphics[trim=0.4in 0in 0.25in 0in,clip,width=1in]{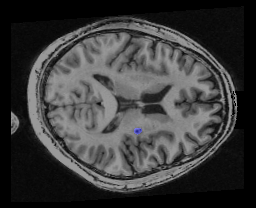} &
	\includegraphics[trim=0.4in 0in 0.25in 0in,clip,width=1in]{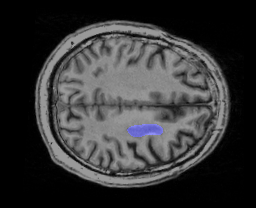} \\ 
	\includegraphics[trim=0in 0in 0in 0in,clip,width=1in]{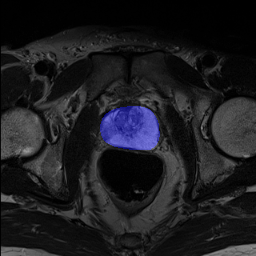} &
	\includegraphics[trim=0in 0in 0in 0in,clip,width=1in]{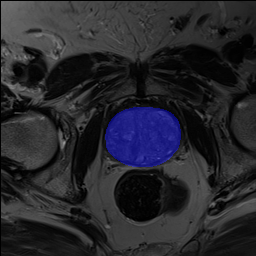} &
	\includegraphics[trim=0.4in 0in 0.25in 0in,clip,width=1in]{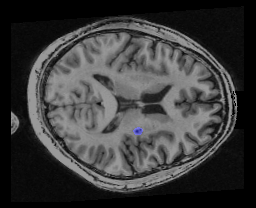} &
	\includegraphics[trim=0.4in 0in 0.25in 0in,clip,width=1in]{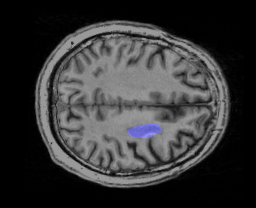} \\ 
	\end{tabular}
	\caption{Ground-truth segmentation (top row) and predicted segmentation results for the proposed approach with setting $\theta_{best}$ + $\alpha$-softmax (middle row) and setting $\theta_{best}$ + $\alpha$-quasimax (bottom row) on the validation set for PROMISE12 (first two columns) and ATLAS (last two columns) datasets.}
	\label{fig:segmentation_demonstration}
\end{figure}

Finally, to visually demonstrate the performance of the proposed approach, qualitative segmentation results are depicted in Fig.~\ref{fig:segmentation_demonstration}. It can be seen that the proposed method achieves good segmentation results for both prostate segmentation task and brain lesion segmentation task.

\section{Conclusion}
This paper described a weakly supervised image segmentation method with tight bounding box supervision. It proposed generalized MIL and smooth maximum approximation to integrate the supervision into the deep neural network. The experiments demonstrate the proposed approach outperforms the state-of-the-art methods. However, there is still performance gap between the weakly supervised approach and the fully supervised method. In the future, it would be interesting to study whether using multi-scale outputs and adding auxiliary object detection task improve the segmentation performance. 



%
%
%
\bibliographystyle{splncs04}
\bibliography{paper142}
\end{document}